\documentclass[runningheads]{llncs}
\usepackage{graphicx}

\usepackage{tikz}
\usepackage{comment}
\usepackage{amsmath,amssymb} 
\usepackage{color}
\usepackage{cite}
\usepackage{pifont}
\newcommand{\cmark}{\ding{52}}%
\newcommand{\xmark}{\ding{56}}%
\usepackage{subcaption}

\usepackage[accsupp]{axessibility}  

\begin{document}
\pagestyle{headings}
\mainmatter
\def\ECCVSubNumber{1846}  

\title{Hierarchical Latent Structure for Multi-Modal Vehicle Trajectory Forecasting} 

\titlerunning{Hierarchical Latent Structure for Trajectory Forecasting}
%
\author{Dooseop Choi\thanks{corresponding author}\inst{1} \and
KyoungWook Min\inst{1}}
\authorrunning{D. Choi and K. Min}
%
\institute{Artificial Intelligence Research Laboratory, ETRI\\
\email{\{d1024.choi,kwmin92\}@etri.re.kr}\\
}
\maketitle

\begin{abstract}
Variational autoencoder (VAE) has widely been utilized for modeling data distributions because it is theoretically elegant, easy to train, and has nice manifold representations. However, when applied to image reconstruction and synthesis tasks, VAE shows the limitation that the generated sample tends to be blurry. We observe that a similar problem, in which the generated trajectory is located between adjacent lanes, often arises in VAE-based trajectory forecasting models. To mitigate this problem, we introduce a hierarchical latent structure into the VAE-based forecasting model. Based on the assumption that the trajectory distribution can be approximated as a mixture of simple distributions (or modes), the low-level latent variable is employed to model each mode of the mixture and the high-level latent variable is employed to represent the weights for the modes. To model each mode accurately, we condition the low-level latent variable using two lane-level context vectors computed in novel ways, one corresponds to vehicle-lane interaction and the other to vehicle-vehicle interaction. The context vectors are also used to model the weights via the proposed mode selection network. To evaluate our forecasting model, we use two large-scale real-world datasets. Experimental results show that our model is not only capable of generating clear multi-modal trajectory distributions but also outperforms the state-of-the-art (SOTA) models in terms of prediction accuracy. Our code is available at https://github.com/d1024choi/HLSTrajForecast.
\end{abstract}

\section{Introduction}
Trajectory forecasting has long been a great interest in autonomous driving since accurate predictions of future trajectories of traffic agents are essential for the safe motion planning of an autonomous vehicle (AV). Many approaches have been proposed for trajectory forecasting in the literature and remarkable progress has been made in recent years. The recent trend in trajectory forecasting is to predict multiple possible trajectories for each agent in the traffic scene. This is because human drivers’ future behavior is uncertain, and consequently, the future motion of the agent naturally exhibits a multi-modal distribution. 

Latent variable models, such as variational autoencoders (VAEs) \cite{Kingma13} and generative adversarial networks (GANs) \cite{Goodfellow}, have been used for modeling the distribution over the agents’ future trajectories. Using latent variables, trajectory forecasting models can learn to capture agent-agent and agent-space interactions from data, and consequently, generate future trajectories that are compliant with the input scene contexts. 

VAEs have been applied in many machine learning applications, including image synthesis \cite{Huang, Vahdat}, language modeling \cite{Bowman, Yang}, and trajectory forecasting \cite{Lee, Casas} because they are theoretically elegant, easy to train, and have nice manifold representations. One of the limitations of VAEs is that the generated sample tends to be blurry (especially in image reconstruction and synthesis tasks) \cite{SZhao}. We found from our experiments that a similar problem often arises in VAE-based trajectory forecasting models. More specifically, it is often found that the generated trajectory is located between adjacent lanes as illustrated in Figure 1. These false positive motion forecasts can cause uncomfortable rides for the AV with plenty of sudden brakes and steering changes \cite{Casas20}. In the rest of this paper, we will refer to this problem as \textit{mode blur} as instance-level lanes are closely related to the modes of the trajectory distribution \cite{Kim}. Mode blur is also found in the recent SOTA model \cite{Cui21} as shown in supplementary materials.

Many approaches have been proposed to mitigate the blurry sample generation problem primarily for image reconstruction or synthesis tasks. In this paper, we introduce a hierarchical latent structure into a VAE-based forecasting model to mitigate mode blur. Based on the assumption that the trajectory distribution can be approximated as a mixture of simple distributions (or modes), the low-level latent variable is employed to model each mode of the mixture and the high-level latent variable is employed to represent the weights for the modes. As a result, the forecasting model is capable of generating clear multi-modal trajectory distributions. To model each mode accurately, we condition the low-level latent variable using two lane-level context vectors (one corresponds to vehicle-lane interaction (VLI) and the other to vehicle-vehicle interaction (V2I)) computed in novel ways. The context vectors are also used to model the weights via the proposed mode selection network. Lastly, we also introduce two techniques to further improve the prediction performance of our model: 1) positional data preprocessing and 2) GAN-based regularization. The preprocessing is introduced based on the fact that vehicles moving along a lane usually try to be parallel to the tangent vector of the lane. The regularization is intended to ensure that the proposed model generates trajectories that match the shape of the lanes well. 

In summary, our contributions are the followings:
\begin{itemize}
    \item[$\bullet$] The hierarchical latent structure is introduced in the VAE-based forecasting model to mitigate mode blur.
    \item[$\bullet$] Two context vectors (one corresponds to the VLI and the other to the V2I) calculated in novel ways are proposed for lane-level scene contexts.
    \item[$\bullet$] Positional data preprocessing and GAN-based regularization are introduced to further improve the prediction performance.
    \item[$\bullet$] Our forecasting model outperforms the SOTA models in terms of prediction accuracy on two large-scale real-world datasets. 
\end{itemize}

\begin{figure}[t]
\centering
\includegraphics[height=5.5cm]{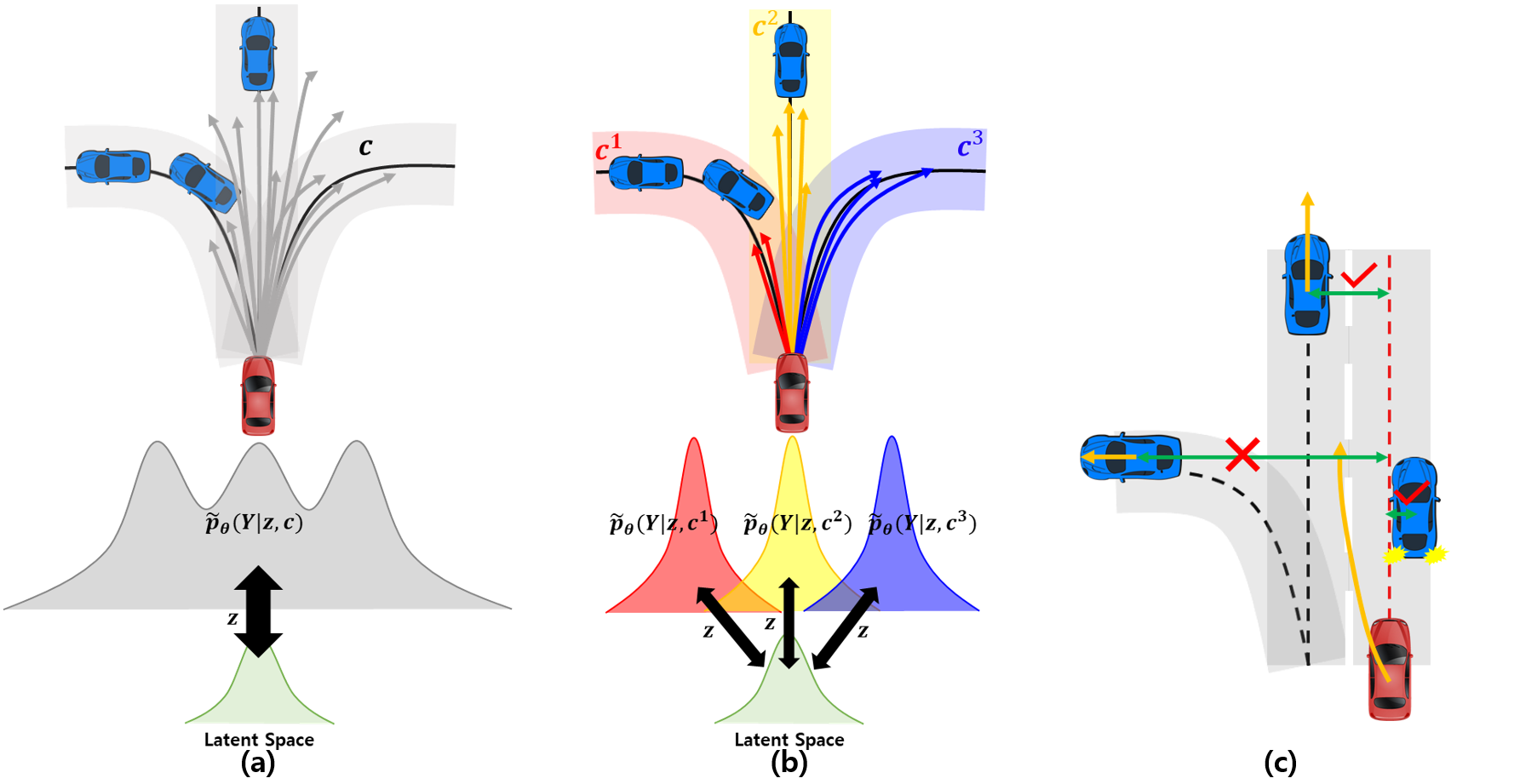}
\caption{Mode blur in trajectory forecasting and our approach. (a) Generated trajectories tend to locate between adjacent lanes. (b) We let a latent variable model each mode of the trajectory distribution to mitigate mode blur. (c) The target vehicle (red) takes into account not only its reference lane (red dashed line) but also the surrounding lanes (black dashed lines) and the surrounding vehicles. Only the surrounding vehicles within a certain distance from the reference lanes (green lines with arrows) influences the future motion of the target vehicle. }
\label{fig1}
\end{figure}

\section{Related Works}
\subsection{Limitations of VAEs} 
The VAE framework has been used to explicitly learn data distributions. The models based on the VAE framework learn mappings from samples in a dataset to points in a latent space and generate plausible samples from variables drawn from the latent space. The VAE-based generative models are known to suffer from two problems: 1) posterior collapse (that the models ignore the latent variable when generating samples) and 2) blurry sample generation. To mitigate the problems, many approaches have been proposed in the literature, primarily for image reconstruction or synthesis tasks \cite{Rezende, Huang, Fu, Kingma, Razavi, Higgins, Zhao, Vahdat}. In trajectory forecasting, some researchers \cite{Casas, Salzmann} have employed the techniques for the mitigation of the posterior collapse. To mitigate the blurry sample generation, \cite{Bhattacharyya} proposed a ``best-of-many" sample objective that leads to accurate and diverse trajectory generation.  

\subsection{Forecasting with Lane Geometry} 
Because the movement of vehicles on the road is greatly restricted by the lane geometry, many works have been proposed to utilize the lane information provided by High-Definition (HD) maps \cite{Cui, Casas, Salzmann, Luo, Fang, Minh, Gao, Liang, Kim, Narayanan}. There are two types of approaches to the representation of the lane information: 1) rasterizing the components of the HD maps on a 2D canvas to obtain the top-view images of the HD maps, 2) representing each component of the HD maps as a series of coordinates of points. In general, Convolutional Neural Network (CNN) is utilized for the former case while Long Short-Term Memory (LSTM) or 1D-CNN is utilized for the latter case to encode the lane information. In this paper, we adopt the second approach. The centerline of each lane in the HD maps is first represented as a series of equally-spaced 2D coordinates and then encoded by an LSTM network. The ability to handle individual lanes in the HD maps allows us to calculate lane-level scene contexts. 

\subsection{Lane-level Scene Context} 
Since instance-level lanes are closely related to the modes of the trajectory distribution, recent works \cite{Luo, Kim, Narayanan, Fang} proposed calculating lane-level scene contexts and using them for generating trajectories. Our work shares the idea with the previous works. However, ours differs from them in the way it calculates the lane-level scene contexts, which leads to significant gains in the prediction performance. Instead of considering only a single lane for a lane-level scene context, we also take into account surrounding lanes along with their relative importance. The relative importance is calculated based on the past motion of the target vehicle, thus reflecting the vehicle-lane interaction. In addition, for the interaction between the target vehicle and surrounding vehicles, we consider only the surrounding vehicles within a certain distance from the reference lane as illustrated in Figure \ref{fig1}c. This approach shows improved prediction performance compared to the existing approaches that consider either all neighbors \cite{Narayanan} or only the most relevant neighbor \cite{Kim}. This result is consistent with the observation that only a subset of surrounding vehicles is indeed relevant when predicting the future trajectory of the target vehicle \cite{Li21}.

\section{Proposed Method}
In this section, we present the details of our trajectory forecasting model.

\subsection{Problem Formulation}
Assume that there are $N$ vehicles in the traffic scene. We aim to generate plausible trajectory distributions $p(\mathbf{Y}_{i}|\mathbf{X}_{i}, \mathcal{C}_{i})$ for the vehicles $\{V_{i}\}_{i=1}^{N}$. Here, $\mathbf{X}_{i}=\mathbf{p}_{i}^{(t-H:t)}$ denotes the positional history of $V_{i}$ for the previous $H$ timesteps at time $t$, $\mathbf{Y}_{i}=\mathbf{p}_{i}^{(t+1:t+T)}$ denotes the future positions of $V_{i}$ for the next $T$ timesteps, and $\mathcal{C}_{i}$ denotes additional scene information available to $V_{i}$. For $\mathcal{C}_{i}$, we use the positional histories of the surrounding vehicles $\{\mathbf{X}_{j}\}_{j=1,j \neq i}^{N}$ and the lane candidates $\mathbf{L}^{(1:M)}$ available for $V_{i}$ at time $t$, where $\mathbf{L}^{m}=\mathbf{l}_{1,...,F}^{m}$ denotes the $F$ equally spaced coordinate points on the centerline of the $m$-th lane. Finally, we note that every positional information is expressed in the coordinate frame defined by $V_{i}$'s current position and heading. According to \cite{Kim}, $p(\mathbf{Y}_{i}|\mathbf{X}_{i}, \mathcal{C}_{i})$ can be re-written as
\begin{equation}
p(\mathbf{Y}_{i}|\mathbf{X}_{i}, \mathcal{C}_{i}) = \sum_{m=1}^{M} \underbrace{p(\mathbf{Y}_{i}|E_{m}, \mathbf{X}_{i}, \mathcal{C}_{i})}_{\text{mode}} \underbrace{p(E_{m} | \mathbf{X}_{i}, \mathcal{C}_{i})}_{\text{weight}},
\label{eqn1}
\end{equation}
where $E_{m}$ denotes the event that $\mathbf{L}^{m}$ becomes the reference lane for $V_{i}$. Equation \ref{eqn1} shows that the trajectory distribution can be expressed as a weighted sum of the distributions which we call $\textit{modes}$. The fact that the modes are usually much simpler than the overall distribution inspired us to model each mode through a latent variable, and sample trajectories from the modes in proportion to their weights as illustrated in Figure \ref{fig1}b.

\subsection{Forecasting Model with Hierarchical Latent Structure}
We introduce two latent variables $\mathbf{z}_{l} \in \mathbb{R}^{D}$ and $\mathbf{z}_{h}\in \mathbb{R}^{M}$ to model the modes and the weights for the modes in Eq. \ref{eqn1}. With the low-level latent variable $\mathbf{z}_{l}$, our forecasting model defines $p(\mathbf{Y}_{i}|E_{m}, \mathbf{X}_{i}, \mathcal{C}_{i})$ by using the decoder network $p_{\theta}(\mathbf{Y}_{i} | \mathbf{z}_{l}, \mathbf{X}_{i}, \mathcal{C}_{i}^{m})$ and the prior network $p_{\gamma}(\mathbf{z}_{l}| \mathbf{X}_{i}, \mathcal{C}_{i}^{m})$ based on 
\begin{equation}
p(\mathbf{Y}_{i}|E_{m}, \mathbf{X}_{i}, \mathcal{C}_{i}) = \int_{\mathbf{z}_{l}} p(\mathbf{Y}_{i} | \mathbf{z}_{l}, \mathbf{X}_{i}, \mathcal{C}_{i}^{m})p(\mathbf{z}_{l}| \mathbf{X}_{i}, \mathcal{C}_{i}^{m})d\mathbf{z}_{l}, \label{eqn0}
\end{equation}
where $\mathcal{C}_{i}^{m} \subset \mathcal{C}_{i}$ denotes the scene information relevant to $\mathbf{L}^{m}$. To train our forecasting model, we employ the conditional VAE framework \cite{Sohn} and optimize the following modified ELBO objective \cite{Higgins}:
\begin{multline}
\mathcal{L}_{ELBO} = -\mathbb{E}_{\mathbf{z}_{l} \sim q_{\phi}}[\log p_{\theta}(\mathbf{Y}_{i} | \mathbf{z}_{l}, \mathbf{X}_{i}, \mathcal{C}_{i}^{m})] \\
+\beta KL(q_{\phi}(\mathbf{z}_{l}| \mathbf{Y}_{i}, \mathbf{X}_{i}, \mathcal{C}_{i}^{m}) || p_{\gamma}(\mathbf{z}_{l}| \mathbf{X}_{i}, \mathcal{C}_{i}^{m})),
\label{eqn2}
\end{multline}
where $\beta$ is a constant and $q_{\phi}(\mathbf{z}_{l}| \mathbf{Y}_{i}, \mathbf{X}_{i}, \mathcal{C}_{i}^{m})$ is the approximated posterior network. The weights for the modes $p(E_{m} | \mathbf{X}_{i}, \mathcal{C}_{i})$ are modeled by the high-level latent variable $\mathbf{z}_{h}$, which is output of the proposed mode selection network $\mathbf{z}_{h} = f_{\varphi}(\mathbf{X}_{i}, \mathcal{C}_{i}^{(1:M)})$.

As shown in Eq. \ref{eqn2} and the definition of the mode selection network, the performance of our forecasting model is dependent on how the lane-level scene information $\mathcal{C}_{i}^{m}$ is utilized along with $\mathbf{X}_{i}$ for defining the lane-level scene context. One can consider two interactions for the lane-level scene context: the VLI and V2I. This is because the future motion of the vehicle is highly restricted not only by the vehicle's motion history but also by the motion histories of the surrounding vehicles and the lane geometry of the road. For the VLI, the existing works \cite{Fang, Kim, Luo, Narayanan} considered only the reference lane. For the V2I, \cite{Kim} considered only one vehicle most relevant to the reference lane, while the others considered all vehicles. In this paper, we present novel ways of defining the two interactions. For the VLI, instead of considering only the reference lane, we also take into account surrounding lanes along with their relative importance, which is calculated based on the target vehicle's motion history. The V2I is encoded through a GNN by considering only surrounding vehicles within a certain distance from the reference lane. Our approach is based on the fact that human drivers often pay attention to surrounding lanes and vehicles occupying the surrounding lanes when driving along the reference lane. Driving behaviors such as lane changes and overtaking are examples. 
\begin{figure}[t]
\centering
\includegraphics[height=4.5cm]{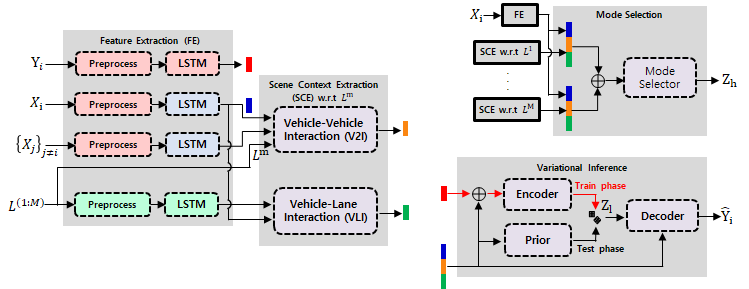}
\caption{Overall architecture of our forecasting model. To generate $K$ future trajectories of $V_{i}$, lane-level scene context vectors $\{\mathbf{c}_{i}^{m}\}_{m=1}^{M}$, each of which corresponds to one of $\mathbf{L}^{(1:M)}$, are first calculated via scene context extraction module. Next, $\{w_{m}\}_{m=1}^{M}$ ($w_{m}$ denotes the probability that $V_{i}$ will drive along $\mathbf{L}^{m}$ in the future) are calculated by using $\mathbf{z}_{h}$. Finally, $\lfloor K \times w_{m} \rfloor$ out of $K$ future trajectories are generated by the decoder network using $\mathbf{c}_{i}^{m}$ and $\mathbf{z}_{l}$.}
\label{fig2}
\end{figure}

\subsection{Proposed Network Structure}
We show in Fig. \ref{fig2} the overall architecture of our forecasting model. In the following sections, we describe the details of our model.

\subsubsection{Feature Extraction Module:}
Three LSTM networks are used to encode the positional data $\{\mathbf{X}_{a}\}_{a=1}^{N}$, $\mathbf{Y}_{i}$, and $\mathbf{L}^{(1:M)}$, respectively. The last hidden state vector of the networks is used for the encoding result. Before the encoding process, we preprocess the positional data. For the vehicles, we calculate the speed and heading at each timestep and concatenate the sequential speed and heading data to the original data along the data dimension. As a result, $\{\mathbf{X}_{a}\}_{a=1}^{N}$ and $\mathbf{Y}_{i}$ have the data dimension of size 4 (x-position, y-position, speed, and heading). For the lanes, at each coordinate point, we calculate the tangent vector and the direction of the tangent vector. The sequential tangential and directional data are concatenated to the original data along the data dimension. As a result, $\mathbf{L}^{(1:M)}$ have the data dimension of size 5 (2D position vector, 2D tangent vector, and direction). We introduce the preprocessing step to make our model better infer the future positions of the target vehicle with the historical speed and heading records and the tangential data, based on that vehicles moving along a lane usually try to be parallel to the tangent vector of the lane. As shown in Table 1, the prediction performance of our model is improved due to the preprocessing step. In the rest of this paper, we use a tilde symbol at the top of a variable to indicate that it is the result of the encoding process. For example, the encoding result of $\mathbf{X}_{i}$ is expressed as $\tilde{\mathbf{X}}_{i}$. 

\subsubsection{Scene Context Extraction Module:} 
Two lane-level context vectors are calculated in this stage. Assume that $\mathbf{L}^{m}$ is the reference lane for $V_{i}$. The context vector $\mathbf{a}_{i}^{m}$ for the VLI is calculated as follows:
\begin{equation}
\mathbf{a}_{i}^{m} = [\tilde{\mathbf{L}}^{m}; \sum_{l=1,l\neq m}^{M} \alpha_{l}\tilde{\mathbf{L}}^{l}],
\label{eqn3}
\end{equation}
where $\{\alpha_{l}\}_{l=1}^{M}$ are the weights calculated through the attention operation \cite{Bahdanau} between $\tilde{\mathbf{X}}_{i}$ and $\tilde{\mathbf{L}}^{(1:M)}$ and the semi-colon denotes the concatenation operation. $\alpha_{l}$ represents the relative importance of the surrounding lane $L^{l}$ compared to the reference lane under the consideration of the past motion of $V_{i}$. As a result, our model can generate plausible trajectories for the vehicles that drive paying attention to multiple lanes. For example, suppose that the vehicle is changing its lane from $L^{m}$ to $L^{l}$. $\alpha_{l}$ will be close to 1 and $\mathbf{a}_{i}^{m}$ can be approximated as $[\tilde{\mathbf{L}}^{m}; \tilde{\mathbf{L}}^{l}]$, thus, our model can generate plausible trajectories corresponding to the lane change. We show in supplementary materials how the target vehicle interacts with the surrounding lanes of the reference lane using some driving scenarios. 

To model the interaction between $V_{i}$ and its surrounding vehicles $\{V_{j}\}_{j \neq i}$, we use a GNN. As we mentioned, only the surrounding vehicles within a certain distance from the reference lane are considered for the interaction; see Fig. \ref{fig1}c. Let $\mathcal{N}_{i}^{m}$ denote the set of the vehicles including $V_{i}$ and its select neighbors. The context vector $\mathbf{b}_{i}^{m}$ for the V2I is calculated as follows: 
\begin{equation} 
\mathbf{m}_{j \to i} = \text{MLP}([\mathbf{p}_{j}^{t}-\mathbf{p}_{i}^{t}; \mathbf{h}_{i}^{k}; \mathbf{h}_{j}^{k}]),
\label{eqn4}
\end{equation}
\begin{equation} 
\mathbf{o}_{i} = \sum_{j \in \mathcal{N}_{i}^{m}, j \neq i} \mathbf{m}_{j \to i},
\label{eqn5}
\end{equation}
\begin{equation} 
\mathbf{h}_{i}^{k+1} = \text{GRU}(\mathbf{o}_{i}, \mathbf{h}_{i}^{k}),
\label{eqn6}
\end{equation}
\begin{equation} 
\mathbf{b}_{i}^{m} = \sum_{j \in \mathcal{N}_{i}^{m}, j \neq i} \mathbf{h}_{j}^{K-1},
\label{eqn7}
\end{equation}
where $\mathbf{h}^{0} = \tilde{\mathbf{X}}$ for all vehicles in $\mathcal{N}_{i}^{m}$. The message passing from $V_{j}$ to $V_{i}$ is defined in Eq. \ref{eqn4} and all messages coming to $V_{i}$ are aggregated by the sum operation as shown in Eq. \ref{eqn5}. After the $K$ rounds of the message passing, the hidden feature vector $\mathbf{h}_{j}^{K-1}$ represents not only the motion history of $V_{j}$ but also the history of the interaction between $V_{j}$ and the others. The distance threshold $\tau$ for $\mathcal{N}_{i}^{m}$ plays the important role in the performance improvement. We explore the choice of $\tau$ value and empirically find that the best performance is achieved with $\tau=5$ meters (the distance between two nearby lane centerlines in straight roads is around 5 meters). Finally, note that we use the zero vector for $\mathbf{b}_{i}^{m}$ when $\mathcal{N}_{i}^{m}$ has the target vehicle only. 

\subsubsection{Mode Selection Network:}
The weights for the modes of the trajectory distribution are calculated by the mode selection network $\mathbf{z}_{h} = f_{\varphi}(\mathbf{X}_{i}, \mathcal{C}_{i}^{(1:M)})$. As instance-level lanes are closely related to the modes, it can be assumed that there are $M$ modes, each corresponding to one of $\mathbf{L}^{(1:M)}$. We calculate the weights from the lane-level scene context vectors $\mathbf{c}_{i}^{m} = [\tilde{\mathbf{X}}_{i}; \mathbf{a}_{i}^{m}; \mathbf{b}_{i}^{m}]$ which condense the information about the modes:
\begin{equation} 
\mathbf{z}_{h} = \text{MLP}_{f_{\varphi}}([\mathbf{c}_{i}^{1};...;\mathbf{c}_{i}^{M}]) \in \mathbb{R}^{M}.
\label{eqn8}
\end{equation}
The softmax operation is applied to $\mathbf{z}_{h}$ to get the final weights $\{w_{m}\}_{m=1}^{M}$. Let $\mathbf{z}_{h}^{SM}$ denote the result of applying the softmax operation to $\mathbf{z}_{h}$. $w_{m}$ is equal to the $m$-th element of $\mathbf{z}_{h}^{SM}$. The lane-level scene context vector is the core feature vector for our encoder, prior, and decoder networks as described in the next section. 

\subsubsection{Encoder, Prior, and Decoder:} 
The approximated posterior $q_{\phi}(\mathbf{z}_{l}| \mathbf{Y}_{i}, \mathbf{X}_{i}, \mathcal{C}_{i}^{m})$, also known as encoder or recognition network, is implemented as MLPs with the encoding of the future trajectory and the lane-level scene context vector as inputs:
\begin{equation} 
\mu_{e}, \sigma_{e} = \text{MLP}_{q_{\phi}}([\tilde{\mathbf{Y}}_{i};\mathbf{c}_{i}^{m}]),
\label{eqn9}
\end{equation}
where $\mu_{e}$ and $\sigma_{e}$ are the mean and standard deviation vectors, respectively. The encoder is utilized in the training phase only because $\mathbf{Y}_{i}$ is not available in the inference phase. The prior $p_{\gamma}(\mathbf{z}_{l}| \mathbf{X}_{i}, \mathcal{C}_{i}^{m})$ is also implemented as MLPs with the context vector as input:
\begin{equation} 
\mu_{p}, \sigma_{p} = \text{MLP}_{p_{\gamma}}(\mathbf{c}_{i}^{m}),
\label{eqn10}
\end{equation}
where $\mu_{p}$ and $\sigma_{p}$ are the mean and standard deviation vectors, respectively. The latent variable $\mathbf{z}_{l}$ is sampled from $(\mu_{e}, \sigma_{e})$ via the re-parameterization trick \cite{Kingma13} during the training and from $(\mu_{p}, \sigma_{p})$ during the inference. 

The decoder network generates the prediction of the future trajectory, $\hat{\mathbf{Y}}_{i}$, via an LSTM network as follows:
\begin{equation} 
\mathbf{e}_{i}^{t} = \text{MLP}_{emb}(\hat{\mathbf{p}}_{i}^{t}),
\label{eqn11}
\end{equation}
\begin{equation} 
\mathbf{h}_{i}^{t+1} = \text{LSTM}([\mathbf{e}_{i}^{t}; \mathbf{c}_{i}^{m}; \mathbf{z}_{l}],\mathbf{h}_{i}^{t}),
\label{eqn12}
\end{equation}
\begin{equation} 
\hat{\mathbf{p}}_{i}^{t+1} = \text{MLP}_{dec}(\mathbf{h}_{i}^{t+1}),
\label{eqn13}
\end{equation}
where we initialize $\hat{\mathbf{p}}_{i}^{0}$ and $\mathbf{h}_{i}^{0}$ as the last observed position of $V_{i}$ and the zero-vector, respectively. 

\subsection{Regularization Through GAN}
To generate more clear image samples, \cite{Larsen} proposed a method that combines VAE and GAN. Based on the observation that the discriminator network implicitly learns a rich similarity metric for images, the typical element-wise reconstruction metric (e.g., $L_{2}$-distance) in the ELBO objective is replaced with a feature-wise metric expressed in the discriminator. In this paper, we also propose training our forecasting model with a discriminator network simultaneously. However, we don't replace the element-wise reconstruction metric with the feature-wise metric since the characteristic of trajectory data is quite different from that of images. We instead use the discriminator to regularize our forecasting model during the training so that the trajectories generated by our model well match the shape of the reference lane. 

The proposed discriminator network is defined as follows:
\begin{equation} 
s = D(\mathbf{Y}_{i}, \mathbf{L}^{m}) = \text{MLP}_{dis}([\tilde{\mathbf{Y}}_{i}; \tilde{\mathbf{L}}^{m}]) \in \mathbb{R}^{1}.
\label{eqn14}
\end{equation}
We explored different choices for the encoding of the inputs to the discriminator network and observed that the following approaches improve the prediction performance: 1) $\tilde{\mathbf{Y}}_{i}$ is the result of encoding $[\mathbf{Y}_{i};\Delta\mathbf{Y}_{i}]$ through an LSTM network where $\Delta\mathbf{Y}_{i} = \Delta \mathbf{p}_{i}^{(t+1:t+T)}$, $\Delta \mathbf{p}_{i}^{t} = \mathbf{p}_{i}^{t} - \mathbf{l}_{f}^{m}$, and $\mathbf{l}_{f}^{m}$ is the coordinate point of $\mathbf{L}^{m}$ closest to $\mathbf{p}_{i}^{t}$, 2) $\tilde{\mathbf{L}}^{m}$ is from the feature extraction module. We also observed that generating trajectories for the GAN objective ($\mathcal{L}_{GAN}$ defined in Eq. \ref{eqn17}) from both the encoder and prior yields better prediction performance, which is consistent with the observations in \cite{Larsen}. However, not back-propagating the error signal from the GAN objective to the encoder and prior does not lead to the performance improvement, which is not consistent with the observations in \cite{Larsen}. 

\subsection{Training Details}
The proposed model is trained by optimizing the following objective:
\begin{equation} 
\mathcal{L} = \mathcal{L}_{ELBO} + \alpha \mathcal{L}_{BCE} + \kappa \mathcal{L}_{GAN}.
\label{eqn15}
\end{equation}
Here, $\mathcal{L}_{BCE}$ is the binary cross entropy loss for the mode selection network and is defined as follows:
\begin{equation} 
\mathcal{L}_{BCE} = \texttt{BCE}(\mathbf{g}^{m}, \texttt{softmax}(\mathbf{z}_{h})),
\label{eqn16}
\end{equation}
where $\mathbf{g}^{m}$ is the one-hot vector indicating the index of the lane, in which the target vehicle traveled in the future timesteps, among the $M$ candidate lanes. $\mathcal{L}_{GAN}$ is the typical adversarial loss defined as follows:
\begin{equation}
\mathcal{L}_{GAN} = \mathbb{E}_{\mathbf{Y}\sim p_{data}}[\mathtt{log} D(\mathbf{Y}, \mathbf{L})] 
+ \mathbb{E}_{\mathbf{z}\sim p_{z}}[\mathtt{log} (1-D(G(\mathbf{z}), \mathbf{L}))], 
\label{eqn17}
\end{equation}
where $G$ denotes our forecasting model. The hyper-parameters ($\alpha$, $\kappa$) in Eq. \ref{eqn15} and $\beta$ in Eq. \ref{eqn2} are set to $1$, $0.01$, and $0.5$, respectively. More details can be found in supplementary materials. 

\subsection{Inference}
Future trajectories for the target vehicle are generated from the modes based on their weights. Assume that $K$ trajectories need to be generated for $V_{i}$. $\lfloor K \times w_{m} \rfloor$ out of $K$ future trajectories are generated by the decoder network using $\mathbf{c}_{i}^{m}$ and $\mathbf{z}_{l}$. In the end, a total of $K$ trajectories can be generated from $\{\mathbf{c}_{i}^{m} \}_{m=1}^{M}$ since $\sum_{m=1}^{M}w_{m}=1$. 

\section{Experiments}
\subsection{Dataset}
Two large-scale real-world datasets, Argoverse Forecasting \cite{Chang} and nuScenes \cite{Caesar}, are used to evaluate the prediction performance of our model. Both provide 2D or 3D annotations of road agents, track IDs of agents, and HD map data. nuScenes includes 1000 scenes, each 20 seconds in length. A 6-second future trajectory is predicted from a 2-second past trajectory for each target vehicle. Argoverse Forecasting is the dataset for the trajectory prediction task. It provides more than 300K scenarios, each 5 seconds in length. A 3-second future trajectory is predicted from a 2-second past trajectory for each target vehicle. Argoverse Forecasting and nuScenes publicly release only training and validation sets. Following the existing works \cite{Kim, Salzmann}, we use the validation set for the test. For the training, we use the training set only.

\subsection{Evaluation Metric}
For the quantitative evaluation of our forecasting model, we employ two popular metrics, average displacement error (ADE) and final displacement error (FDE), defined as follows:
\begin{equation}
ADE(\hat{\mathbf{Y}}, \mathbf{Y}) = \frac{1}{T} \sum_{t=1}^{T} || \hat{\mathbf{p}}^{t} - \mathbf{p}^{t}||_{2},
\label{eqn18}
\end{equation}
\begin{equation}
FDE(\hat{\mathbf{Y}}, \mathbf{Y}) = || \hat{\mathbf{p}}^{T} - \mathbf{p}^{T}||_{2},
\label{eqn19}
\end{equation}
where $\mathbf{Y}$ and $\hat{\mathbf{Y}}$ respectively denote the ground-truth trajectory and its prediction. In the rest of this paper, we denote $ADE_{K}$ and $FDE_{K}$ as the minimum of ADE and FDE among the $K$ generated trajectories, respectively. It is worth noting that $ADE_{1}$ and $FDE_{1}$ metrics shown in the tables presented in the later sections represent the average quality of the trajectories generated for $\mathbf{Y}$. Our derivation can be found in the supplementary materials. On the other hand, $ADE_{K}$ and $FDE_{K}$ represent the quality of the trajectory closest to the ground-truth among the $K$ generated trajectories. We will call $ADE_{K\ge 12}$ and $FDE_{K\ge 12}$ metrics in the tables the \textit{best quality} in the rest of this paper. According to \cite{Casas}, the average quality and the best quality are complementary and evaluate the precision and coverage of the predicted trajectory distributions, respectively.

\begin{table}[t]
\scriptsize

\begin{subtable}{1.0\linewidth} \centering
{
\begin{tabular}{|c|c c c c|c|c|}
\hline
Model & PDP & VLI & V2I & GAN & $ADE_{1}$/$FDE_{1}$ & $ADE_{15}$/$FDE_{15}$ \\
\hline
\textbf{M1} & \xmark & \xmark & \xmark & \xmark & 3.15/7.53 & 0.95/1.82\\ 
\textbf{M2} & \cmark & \xmark & \xmark & \xmark & 3.03/7.22 & 0.93/1.80\\ 
\textbf{M3} & \cmark & \cmark & \xmark & \xmark & 2.91/7.00 & 0.94/1.82\\ 
\textbf{M4} & \cmark & \cmark & \cmark & \xmark & 2.67/6.38 & 0.91/1.77\\ 
\textbf{M5} & \cmark & \cmark & \cmark & \cmark & \textbf{2.64}/\textbf{6.32} & \textbf{0.89}/\textbf{1.72}\\ 
\hline
\end{tabular}}
\caption{}\label{table:one-a}
\end{subtable}

\begin{subtable} {1.0\linewidth}\centering
{
\begin{tabular}{|c|c|c|}
\hline
Model & $ADE_{1}$/$FDE_{1}$ & $ADE_{15}$/$FDE_{15}$ \\
\hline
\textbf{Ours} ($\tau$=1) & 2.66/\textbf{6.31} & 0.92/1.76\\
\textbf{Ours} ($\tau$=5) & \textbf{2.64}/6.32 &  \textbf{0.89}/\textbf{1.72}\\
\textbf{Ours} ($\tau$=10) & 2.65/6.34 & 0.95/1.84\\
\textbf{Ours}+\textbf{All}  & 2.67/6.34 & 1.00/1.98\\
\textbf{Ours}+\textbf{Rel}  & 2.75/6.52 & 0.91/1.77\\
\hline
\end{tabular}}
\caption{}\label{table:one-b}
\end{subtable}

\caption{Ablation study conducted on nuScenes} \label{table:one}
\end{table}

\begin{figure}[t]
\centering
\includegraphics[height=2.0cm]{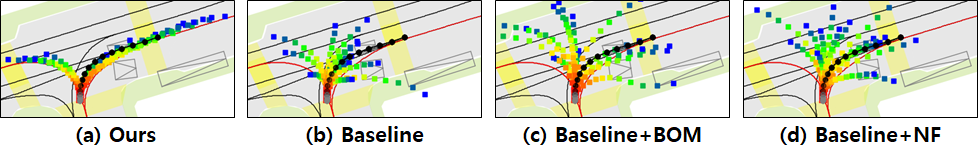}
\caption{Mode blur example}
\label{fig3}
\end{figure}

\begin{table}[t]
\scriptsize
\begin{center}
\begin{tabular}{|c|c|c|c|c|c|c|c|c|}
\hline
Model & $ADE_{1}$ & $FDE_{1}$ & $ADE_{5}$ & $FDE_{5}$ & $ADE_{10}$ & $FDE_{10}$ & $ADE_{15}$ & $FDE_{15}$\\
\hline
CoverNet \cite{Minh}     & 3.87 & 9.26 & 1.96 &  - & 1.48 & -  & -  & -  \\
Trajectron++ \cite{Salzmann} & - & 9.52 & 1.88 & - & 1.51 & - & - & - \\
AgentFormer \cite{Yuan_iccv21} & - & - & 1.86 & 3.89 & 1.45 & 2.86 & - & - \\
ALAN \cite{Narayanan}         & 4.67 & 10.0 & 1.77 & \underline{3.32} & \underline{1.10} & \textbf{1.66} & - & -\\
LaPred \cite{Kim}       & \underline{3.51} & \underline{8.12} & \underline{1.53} & 3.37 & 1.12 & 2.39 & 1.10 & 2.34\\
MHA-JAM \cite{Messaoud}     & 3.69 & 8.57 & 1.81 & 3.72 & 1.24 & 2.21 & \underline{1.03} & \textbf{1.7} \\
\hline
Ours           & $\textbf{2.64}_{0.87\downarrow}$ & $\textbf{6.32}_{1.8\downarrow}$ & $\textbf{1.33}_{0.2\downarrow}$ & $\textbf{2.92}_{0.4\downarrow}$ & $\textbf{1.04}_{0.06\downarrow}$ & $\underline{2.15}_{0.49\uparrow}$ & $\textbf{0.89}_{0.14\downarrow}$ & $\underline{1.72}_{0.02\uparrow}$\\
\hline
\end{tabular}
\end{center}
\caption{Quantitative comparison on nuScenes}

\scriptsize
\begin{center}
\begin{tabular}{|c|c|c|c|c|c|c|c|c|}
\hline
Model & $ADE_{1}$ & $FDE_{1}$ & $ADE_{5}$ & $FDE_{5}$ & $ADE_{6}$ & $FDE_{6}$ & $ADE_{12}$ & $FDE_{12}$\\
\hline
DESIRE \cite{Lee} & 2.38 & 4.64 & 1.17 & 2.06 & 1.09 & 1.89 & 0.90 & 1.45 \\
R2P2 \cite{Rhinehart} & 3.02 & 5.41 & 1.49 & 2.54 & 1.40 & 2.35 & 1.11 & 1.77 \\
VectorNet \cite{Gao} & 1.66 & 3.67 & - & - & - & - & - & - \\
LaneAttention \cite{Luo} & \underline{1.46} & \underline{3.27} & - & - & 1.05 & 2.06 & - & - \\
LaPred \cite{Kim} & 1.48 & 3.29 & \underline{0.76} & \underline{1.55} & \underline{0.71} & \underline{1.44} & \underline{0.60} & \underline{1.15} \\
\hline
Ours & $\textbf{1.44}_{0.02\downarrow}$ & $\textbf{3.15}_{0.12\downarrow}$ & $\textbf{0.70}_{0.06\downarrow}$ & $\textbf{1.35}_{0.2\downarrow}$ & $\textbf{0.65}_{0.06 \downarrow}$ & $\textbf{1.24}_{0.2\downarrow}$ & $\textbf{0.51}_{0.09\downarrow}$ & $\textbf{0.85}_{0.3\downarrow}$ \\
\hline
\end{tabular}
\end{center}
\caption{Quantitative comparison on Argoverse Forecasting}
\end{table}

\subsection{Ablation Study}
\subsubsection{Performance Gain over Baseline}
In Table \ref{table:one-a}, we present the contributions of each idea to the performance gain over a baseline. \textbf{M1} denotes the baseline that does not use the positional data preprocessing (PDP), VLI, V2I, and GAN regularization proposed in this paper. We can see from the table that the average quality of the generated trajectories is improved by both the PDP and the VLI (\textbf{M1} v.s. \textbf{M2} v.s. \textbf{M3}). The improvement due to the VLI is consistent with the observation in \cite{Kim} that consideration of multiple lane candidates is more helpful than using a single best lane candidate in predicting the future trajectory. Both the average quality and the best quality are much improved by the V2I (\textbf{M3} v.s. \textbf{M4}). The accurate trajectory prediction for the vehicles waiting for traffic lights is the most representative case of the performance improvement by the V2I. Due to the past movement of the neighboring vehicles waiting for the traffic light, our model can easily conclude that the target vehicle will also be waiting for the traffic light. Finally, the prediction performance is further improved by the GAN regularization (\textbf{M4} v.s. \textbf{M5}). As seen in Eq. \ref{eqn14}, our discriminator uses a future trajectory along with the reference lane to discriminate between fake trajectories and real trajectories.

\subsubsection{Effect of Surrounding Vehicle Selection Mechanism}
In Table \ref{table:one-b}, we show the effect of the surrounding vehicle selection mechanism on the prediction performance of our model. Here, \textbf{Ours} ($\tau$) denotes our model in which only the surrounding vehicles within $\tau$ meters from the reference lane are considered. \textbf{Ours}+\textbf{Rel} and \textbf{Ours}+\textbf{All} denote our model in which the most relevant vehicle and all the vehicles are considered, respectively. We can see from the table that \textbf{Ours} with $\tau=5$ shows the best performance. This result demonstrates that considering only surrounding vehicles within a certain distance from the reference lane is effective in modeling the V2I from a lane-level perspective.

\subsubsection{Hierarchical Latent Structure}
We show in Fig. \ref{fig3} the generated trajectories for a particular scenario to demonstrate how helpful the introduction of the hierarchical latent structure would be for the mitigation of mode blur. In the figure, \textbf{Baseline} denotes the VAE-based forecasting model in which a latent variable is trained to model the trajectory distribution. \textbf{Baseline}+\textbf{BOM} and \textbf{Baseline}+\textbf{NF} respectively denote \textbf{Baseline} trained with the best-of-many (BOM) sample objective \cite{Bhattacharyya} and normalizing flows (NF) \cite{Rezende}. We introduce NF since the blurry sample generation is often attributed to the limited capability of the approximated posterior \cite{Huang} and NF is a powerful framework for building flexible approximated posterior distributions \cite{Kingma}. In the figure, gray and black circles indicate historical and future positions, respectively. Squares with colors indicate the predictions of the future positions. Time is encoded in the rainbow color map ranging from red (0s) to blue (6s). Red solid lines indicate the centerlines of the candidate lanes. For the scenario, fifteen trajectories were generated. We can see in the figure that the proposed model generates trajectories that are aligned with the lane candidates. In contrast, neither normalizing flows nor BOM objective can help a lot for the mitigation of mode blur.
\begin{figure}[t]
\centering
\includegraphics[height=10.0cm]{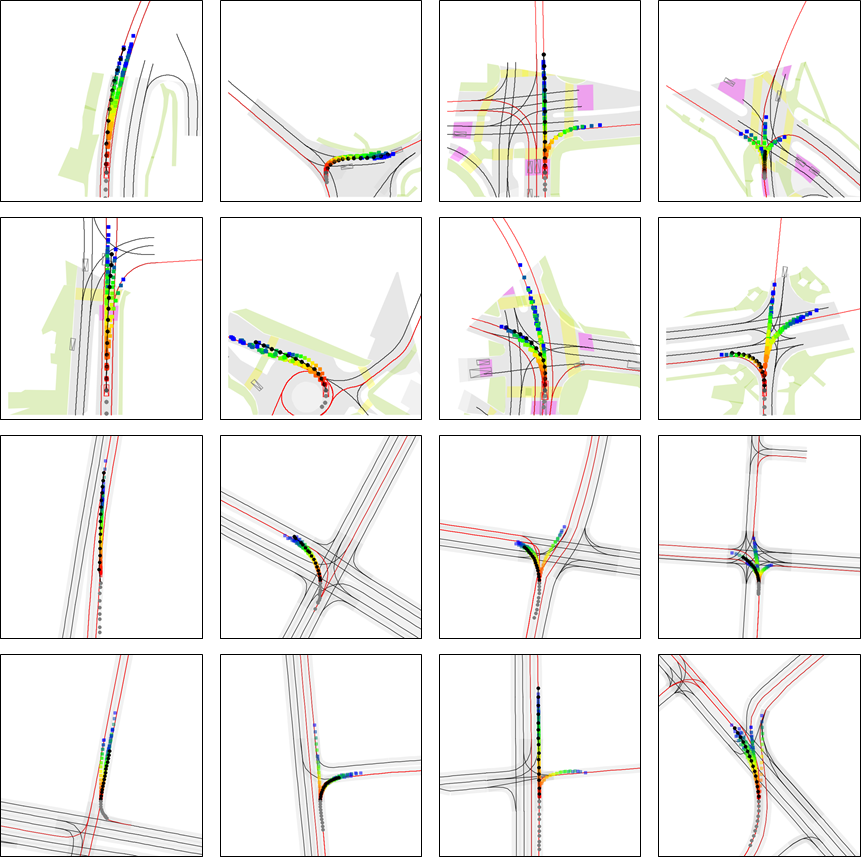}
\caption{Trajectory prediction examples of our forecasting model on nuScenes (the first and second rows) and Argoverse Forecasting (the third and fourth rows)}
\label{fig4}
\end{figure}

\subsection{Performance Evaluation}
\subsubsection{Quantitative Evaluation}
We compare our forecasting model with the existing models objectively. The results are shown in Table 2 and 3. Note that the bold and underline indicate the best and second-best performance, respectively. The values in the subscript indicate the performance gain over the second-best or loss over the best. Finally, the values in the table are from the corresponding papers and \cite{Kim}. Table 2 presents the results on Nuscenes. It shows that our model outperforms the SOTA models \cite{Narayanan, Kim, Yuan_iccv21} on most of the metrics. In particular, the performance gains over the SOTA models in the $ADE_{K \leqq 5}$ and $FDE_{K \leqq 5}$ metrics are significant. Consequently, it can be said that the trajectories generated from our model, on average, are more accurate than those from the SOTA models. On the other hand, \cite{Narayanan} shows the significant performance on $FDE_{10}$. This is because, in \cite{Narayanan}, the vehicle trajectory is defined along the centerlines in a 2D curvilinear normal-tangential coordinate frame, so that the predicted trajectory is well aligned with the centerlines. However, \cite{Narayanan} shows the poorest performance in the average quality. Table 3 presents the results on Argoverse Forecasting. It is seen that our forecasting model outperforms the SOTA models \cite{Kim, Luo} on all the metrics. The $ADE_{12}$ and $FDE_{12}$ results show that our model achieves much better performance in the best quality compared to the models. However, the performance gain over the second-best model in the average quality is not significant. In short, our forecasting model exhibits remarkable performance in the average and best quality on the two large-scale real-world datasets.

\subsubsection{Qualitative Evaluation}
Figure \ref{fig4} illustrates the trajectories generated by our model for particular scenarios in the test dataset. Note that fifteen and twelve trajectories were generated for each scenario in nuScenes and Argoverse Forecasting, respectively. We can see in the figure that the generated trajectories are well distributed along admissible routes. In addition, the shape of the generated trajectory matches the shape of the candidate lane well. These results verify that the trajectory distribution is nicely modeled by the two latent variables conditioned by the proposed lane-level scene context vectors. It is noticeable that our model can generate plausible trajectories for the driving behaviors that require simultaneous consideration of multiple lanes. The first and third figures in the first column show the scenario where the target vehicle has just started changing lanes, and the second shows the scenario where the target vehicle is in the middle of a lane change. For both scenarios, our model generates plausible trajectories corresponding to both changing lanes and returning back to its lane. Finally, the last figure in the first column shows the scenario where the target vehicle is in the middle of a right turn. Our model well captures the motion ambiguity of the vehicle that can keep a lane or change lanes. 

\section{Conclusions}
In this paper, we proposed a VAE-based trajectory forecasting model that exploits the hierarchical latent structure. The hierarchy in the latent space was introduced to the forecasting model to mitigate mode blur by modeling the modes of the trajectory distribution and the weights for the modes separately. For the accurate modeling of the modes and weights, we introduced two lane-level context vectors calculated in novel ways, one corresponds to the VLI and the other to the V2I. The prediction performance of the model was further improved by the two techniques, positional data preprocessing and GAN-based regularization, introduced in this paper. Our experiments on two large-scale real-world datasets demonstrated that the model is not only capable of generating clear multi-modal trajectory distributions but also outperforms the SOTA models in terms of prediction accuracy.

\textbf{Acknowledgment} This research work was supported by the Institute of
Information $\&$ Communications Technology Planning $\&$ Evaluation (IITP) grant funded by the Korean government (MSIP) (No. 2020-0-00002, Development of standard SW platform-based autonomous driving technology to solve social problems of mobility and safety for public
transport-marginalized communities)


\clearpage
%
%
\bibliographystyle{splncs04}
\bibliography{main}

\begin{thebibliography}{10}
\providecommand{\url}[1]{\texttt{#1}}
\providecommand{\urlprefix}{URL }
\providecommand{\doi}[1]{https://doi.org/#1}

\bibitem{Bahdanau}
Bahdanau, D., Cho, K., Bengio, Y.: Neural machine translation by jointly
  learning to align and translate. In: Int. Conf. on Learn. Represent. (2015)

\bibitem{Bhattacharyya}
Bhattacharyya, A., Schiele, B., Fritz, M.: Accurate and diverse sampling of
  sequences based on a best-of-many sample objective. In: IEEE Conf. Comput.
  Vis. Pattern Recog. (2018)

\bibitem{Bowman}
Bowman, S.R., Vilnis, L., Vinyals, O., Dai, A.M., Jozefowicz, R., Bengio, S.:
  Generating sentences from a continuous space. In: arXiv:1511.06349 (2015)

\bibitem{Caesar}
Caesar, H., Bankiti, V., Lang, A.H., Vora, S., Liong, V.E., Xu, Q., Krishnan,
  A., Pan, Y., Baldan, G., Beijbom, O.: nuscenes: a multimodal dataset for
  autonomous driving. In: IEEE Conf. Comput. Vis. Pattern Recog. (2020)

\bibitem{Casas}
Casas, S., Gulino, C., Suo, S., Luo, K., Liao, R., Urtasun, R.: Implicit latent
  variable model for scene-consistent motion forecasting. In: Eur. Conf.
  Comput. Vis. (2020)

\bibitem{Casas20}
Casas, S., Gulino, C., Suo, S., Urtasun, R.: The importance of prior knowledge
  in precise multimodal prediction. In: Int. Conf. Intell. Robots Syst. (2020)

\bibitem{Chang}
Chang, M.F., Lambert, J., Sangkloy, P., Singh, J., Bak, S., Hartnett, A., Wang,
  D., Carr, P., Lucey, S., Ramanan, D., Hays, J.: Argoverse: 3d tracking and
  forecasting with rich maps. In: IEEE Conf. Comput. Vis. Pattern Recog. (2019)

\bibitem{Cui21}
Cui, A., Sadat, A., Casas, S., Liao, R., Urtasun, R.: Lookout: diverse
  multi-future prediction and planning for self-driving. In: Int. Conf. Comput.
  Vis. (2021)

\bibitem{Cui}
Cui, H., Radosavljevic, V., F.-C.Chou, Lin, T.H., Nguyen, T., Huang, T.K.,
  Schneider, J., Djuric, N.: Multimodal trajectory predictions for autonomous
  driving using deep convolutional networks. In: IEEE Int. Conf. Robotics and
  Automation (2019)

\bibitem{Fang}
Fang, L., Jiang, Q., Shi, J., Zhou, B.: Tpnet: trajectory proposal network for
  motion prediction. In: IEEE Conf. Comput. Vis. Pattern Recog. (2020)

\bibitem{Fu}
Fu, H., Li, C., Liu, X., Gao, J., Celikyilmaz, A., Carin, L.: Cyclical
  annealing schedule: A simple approach to mitigating kl vanishing. In: NAACL
  (2019)

\bibitem{Gao}
Gao, J., Sun, C., Zhao, H., Shen, Y., Anguelov, D., Li, C., Schmid, C.:
  Vectornet: encoding hd maps and agent dynamics from vectorized
  representation. In: IEEE Conf. Comput. Vis. Pattern Recog. (2020)

\bibitem{Goodfellow}
Goodfellow, I., Abadie, J.P., Mirza, M., Xu, B., Farley, D.W., Ozair, S.,
  Courville, A., Bengio, Y.: Generative adversarial nets. In: Adv. Neural
  Inform. Process. Syst. (2014)

\bibitem{Higgins}
Higgins, I., Matthey, L., Pal, A., Burgess, C., Glorot, X., Botvinick, M.,
  Mohamed, S., Lerchner, A.: beta-vae: learning basic visual concepts with a
  constrained variational framework. In: Int. Conf. on Learn. Represent. (2017)

\bibitem{Huang}
Huang, H., Li, Z., He, R., Sun, Z., Tan, T.: Introvae: Introspective
  variational autoencoders for photographic image synthesis. In: Adv. Neural
  Inform. Process. Syst. (2018)

\bibitem{Kim}
Kim, B., Park, S.H., Lee, S., Khoshimjonov, E., Kum, D., Kim, J., Kim, J.S.,
  Choi, J.W.: Lapred: lane-aware prediction of multi-modal future trajectories
  of dynamic agents. In: IEEE Conf. Comput. Vis. Pattern Recog. (2021)

\bibitem{Kingma15}
Kingma, D.P., Ba, L.J.: Adam: a method for stochastic optimization. In: Int.
  Conf. on Learn. Represent. (2015)

\bibitem{Kingma}
Kingma, D.P., Salimans, T., Jozefowicz, R., Chen, X., Sutskever, I., Welling,
  M.: Improved variational inference with inverse autoregressive flow. In: Adv.
  Neural Inform. Process. Syst. (2016)

\bibitem{Kingma13}
Kingma, D.P., Welling, M.: Auto-encoding variational bayes. In: arXiv:1312.6114
  (2013)

\bibitem{Larsen}
Larsen, A.B.L., Sonderby, S.K., Larochelle, H., Winther, O.: Autoencoding
  beyond pixels using a learned similarity metric. In: Int. Conf. on Learn.
  Represent. (2016)

\bibitem{Lee}
Lee, N., Choi, W., Vernaza, P., Choy, C.B., Torr, P.H.S., Chan, M.: Desire:
  Distant future prediction in dynamic scenes with interacting agents. In: IEEE
  Conf. Comput. Vis. Pattern Recog. (2017)

\bibitem{Li21}
Li, J., Yang, F., Ma, H., Malla, S., Tomizuka, M., Choi, C.: Rain: reinforced
  hybrid attention inference network for motion forecasting. In: Int. Conf.
  Comput. Vis. (2021)

\bibitem{Liang}
Liang, M., Yang, B., Hu, R., Chen, Y., Liao, R., Feng, S., Urtasun, R.:
  Learning lane graph representations for motion forecasting. In: Eur. Conf.
  Comput. Vis. (2020)

\bibitem{Luo}
Luo, C., Sun, L., Dabiri, D., Yuille, A.: Probabilistic multi-modal trajectory
  prediction with lane attention for autonomous vehicles. In: IEEE Conf.
  Intell. Robots Syst. (2020)

\bibitem{Messaoud}
Messaoud, K., Deo, N., Trivedi, M.M., Nashashibi, F.: Trajectory prediction for
  autonomous driving based on multi-head attention with joint agent-map
  representation. In: arXiv:2005.02545 (2020)

\bibitem{Narayanan}
Narayanan, S., Moslemi, R., Pittaluga, F., Liu, B., Chandraker, M.:
  Divide-and-conquer for lane-aware diverse trajectory prediction. In: IEEE
  Conf. Comput. Vis. Pattern Recog. (2021)

\bibitem{Minh}
P-Minh, T., Grigore, E.C., Boulton, F.A., Beijbom, O., Wolff, E.M.: Covernet:
  multimodal behavior prediction using trajectory sets. In: IEEE Conf. Comput.
  Vis. Pattern Recog. (2020)

\bibitem{Razavi}
Razavi, A., Oord, A., Poole, B., Vinyals, O.: Preventing posterior collapse
  with delta-vaes. In: Int. Conf. on Learn. Represent. (2019)

\bibitem{Rezende}
Rezende, D.J., Mohamad, S.: Variational inference with normalizing flows. In:
  Int. Conf. on Mach. Learn. (2015)

\bibitem{Rhinehart}
Rhinehart, N., Kitani, K.M., Vernaza, P.: R2p2: a reparameterized pushforward
  policy for diverse, precise generative path forecasting. In: Eur. Conf.
  Comput. Vis. (2018)

\bibitem{Salzmann}
Salzmann, T., Ivanovic, B., Chakravarty, P., Pavone, M.: Trajectron++:
  dynamically-feasible trajectory forecasting with heterogeneous data. In: Eur.
  Conf. Comput. Vis. (2020)

\bibitem{Sohn}
Sohn, K., Lee, H., Yan, X.: Learning structured output representation using
  deep conditional generative models. In: Adv. Neural Inform. Process. Syst.
  (2015)

\bibitem{Vahdat}
Vahdat, A., Kautz, J.: Nvae: a deep hierarchical variational autoencoder. In:
  Adv. Neural Inform. Process. Syst. (2020)

\bibitem{Yang}
Yang, Z., Hu, Z., Salakhutdinov, R., B.-Kirkpatrick, T.: Improved variational
  autoencoders for text modeling using dilated convolutions. In: Int. Conf. on
  Mach. Learn. (2017)

\bibitem{Yuan_iccv21}
Yuan, Y., Weng, X., Ou, Y., Kitani, K.: Agentformer: agent-aware transformers
  for socio-temporal multi-agent forecasting. In: arXiv:2103.14023 (2021)

\bibitem{Zhao}
Zhao, S., Song, J., Ermon, S.: Infovae: information maximizing variational
  autoencoders. In: arXiv:1706.02262 (2017)

\bibitem{SZhao}
Zhao, S., Song, J., Ermon, S.: Towards a deeper understanding of variational
  autoencoding models. In: arXiv:1702.08658v1 (2017)

\end{thebibliography}

\vfill\pagebreak
\appendix
\section{Visualization of Vehicle-Lane Interaction (VLI)}
\begin{figure}[t]
\centering
\includegraphics[height=7.5cm]{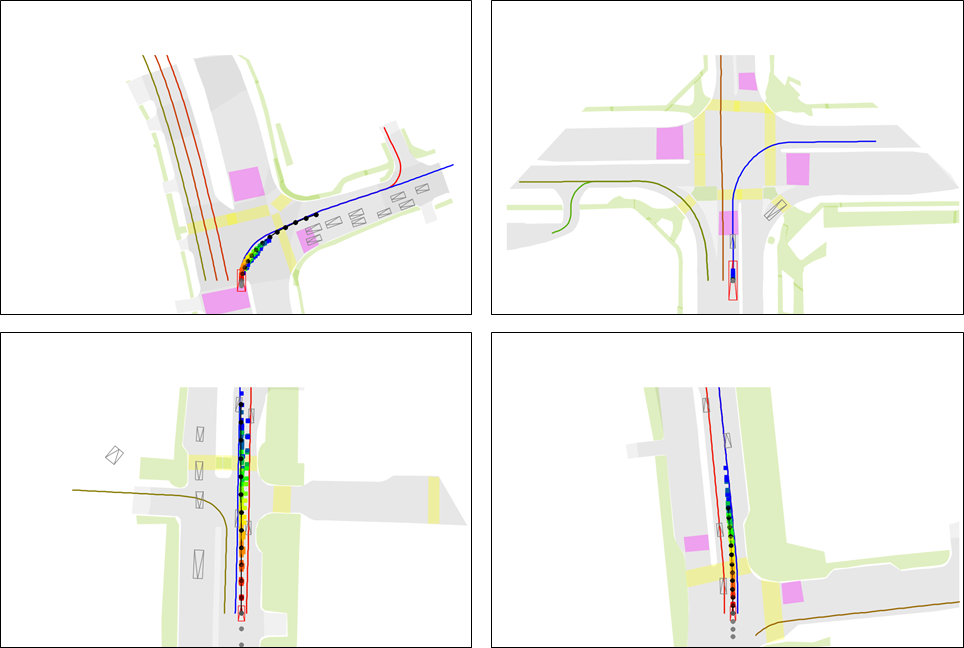}
\caption{VLI visualization}
\label{figA1}
\end{figure}

As we mentioned in the paper, for the calculation of the lane-level context vector $\mathbf{a}_{i}^{m}$, we use not only the reference lane but also the surrounding lanes with their relative importance. This idea is based on the fact that human drivers often pay attention to surrounding lanes when driving along the reference lane. To show how our model pays attention to the surrounding lanes for the target vehicle, we use four scenarios in nuScenes and show the results in Figure \ref{figA1}. In the figure, blue lines denote the reference lanes while the others denote the surrounding lanes. The surrounding lanes of high importance are shown in red and the surrounding lanes of low importance are shown in green. We can see in the figure that our forecasting model pays more attention to the surrounding lanes that are close to the reference lane.

\section{Mode Blur in SOTA Model}
\begin{figure}[t]
\centering
\includegraphics[height=3.5cm]{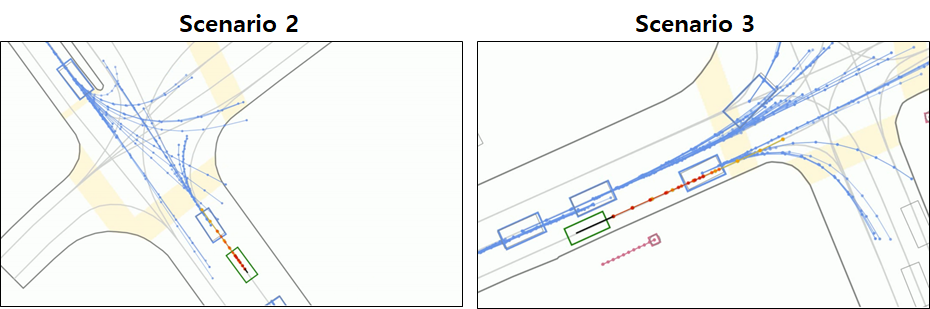}
\caption{Trajectory prediction examples of \cite{Cui21}}
\label{figA2}
\end{figure}

We show in Figure \ref{figA2} the prediction examples of the state-of-the-art model \cite{Cui21}. We note here that the figure is identical to the figure illustrated in the supplementary material of \cite{Cui21}. The model is built upon \cite{Casas}, which is based on the VAE framework and learns a diverse joint distribution over multi-agent future trajectories in a traffic scene. In the figure, green and light blue bounding boxes respectively denote the AV and surrounding vehicles. The solid lines with light blue dots denote the predicted trajectories for the surrounding vehicles. We can see in the figure that some trajectories are located between adjacent lanes, which can cause uncomfortable rides for the AV with plenty of sudden brakes and steering changes \cite{Casas20}.

\section{Further Explanation to Average Quality}
We mentioned in the paper that $ADE_{1}$ and $FDE_{1}$ metrics shown in the tables presented in the paper represent the average quality of the trajectories generated for the ground-truth trajectory $\mathbf{Y}$. The $ADE_{1}$ metric in the table is calculated as
\begin{equation}
\begin{aligned}
ADE_{1} &= \frac{1}{|\mathcal{D}|} \sum_{\mathbf{Y} \in \mathcal{D}} ADE(\hat{\mathbf{Y}}, \mathbf{Y}),
\end{aligned}
\end{equation}
where $\mathcal{D}$ is the test dataset and $\hat{\mathbf{Y}}$ is the prediction of $\mathbf{Y}$. Because there are relatively few distinct actions that can be taken by a vehicle over a reasonable time horizon (3 to 6 seconds) \cite{Minh}, the ground-truth trajectories in $\mathcal{D}$ can be clustered into multiple groups, where the trajectories of each group are very close to each other in Euclidean space. Assume that there are $N$ groups in $\mathcal{D}$ and let $\mathcal{Y}_{i}$ denote the $i$-th group. Then Eqn. 1 can be expressed as
\begin{equation}
\begin{aligned}
ADE_{1} &=\frac{1}{|\mathcal{D}|} \{ \sum_{\mathbf{Y} \in \mathcal{Y}_{1}} ADE(\hat{\mathbf{Y}}, \mathbf{Y})+...+\sum_{\mathbf{Y} \in \mathcal{Y}_{N}} ADE(\hat{\mathbf{Y}}, \mathbf{Y}) \}
\\
&=\frac{|\mathcal{Y}_{1}|}{|\mathcal{D}|} \frac{1}{|\mathcal{Y}_{1}|} \sum_{\mathbf{Y} \in \mathcal{Y}_{1}} ADE(\hat{\mathbf{Y}}, \mathbf{Y})+...+\frac{|\mathcal{Y}_{N}|}{|\mathcal{D}|} \frac{1}{|\mathcal{Y}_{N}|} \sum_{\mathbf{Y} \in \mathcal{Y}_{N}} ADE(\hat{\mathbf{Y}},
\mathbf{Y})
\\
&= w_{1} \frac{1}{|\mathcal{Y}_{1}|} \sum_{\mathbf{Y} \in \mathcal{Y}_{1}} ADE(\hat{\mathbf{Y}}, \mathbf{Y})+...+w_{N} \frac{1}{|\mathcal{Y}_{N}|} \sum_{\mathbf{Y} \in \mathcal{Y}_{N}} ADE(\hat{\mathbf{Y}},\mathbf{Y})
\\
&= w_{1} AADE(\mathcal{Y}_{1}) + ... + w_{N} AADE(\mathcal{Y}_{N}),
\end{aligned}
\end{equation}
where $\sum_{i=1}^{N}w_{i}=1$. Since the trajectories of each group are very close to each other in Euclidean space, $AADE(\mathcal{Y}_{i})$ in the last line of Eqn. 2 can be approximated as 
\begin{equation}
AADE(\mathcal{Y}_{i}) = \frac{1}{|\mathcal{Y}_{i}|} \sum_{\mathbf{Y} \in \mathcal{Y}_{i}} ADE(\hat{\mathbf{Y}}, \mathbf{Y}) \approx \frac{1}{K} \sum_{k=1}^{K} ADE(\hat{\mathbf{Y}}_{k}, \mathbf{Y}_{r})
\end{equation}
where $K=|\mathcal{Y}_{i}|$ is large enough. Here $\mathbf{Y}_{r}$ and $\hat{\mathbf{Y}}_{k}$ are the most representative trajectory in $\mathcal{Y}_{i}$ and its $k$-th prediction, respectively. The last term of Eqn. 3 is the average quality of the $K$ trajectories generated for $\mathbf{Y}_{r}$. Consequently, the $ADE_{1}$ metric represents the average quality. The same derivation can be applied for the $FDE_{1}$ metric.
\begin{table}[t]
\scriptsize

\begin{subtable} {1.0\linewidth}\centering
{
\begin{tabular}{|c|c|c|}
\hline
Model & $ADE_{1}$/$FDE_{1}$ & $ADE_{15}$/$FDE_{15}$ \\
\hline
\textbf{Ours}+\textbf{Multi}  & 2.64/6.32 &  0.89/1.72\\
\textbf{Ours}+\textbf{Single}  & 2.64/6.32 &  0.97/1.95\\

\hline
\end{tabular}}
\caption{nuScenes}\label{tableA:one-a}
\end{subtable}

\begin{subtable} {1.0\linewidth}\centering
{
\begin{tabular}{|c|c|c|}
\hline
Model & $ADE_{1}$/$FDE_{1}$ & $ADE_{12}$/$FDE_{12}$ \\
\hline
\textbf{Ours}+\textbf{Multi}  & 1.44/3.15 &  0.51/0.85\\
\textbf{Ours}+\textbf{Single}  & 1.44/3.16 &  0.53/0.92\\
\hline
\end{tabular}}
\caption{Argoverse Forecasting}\label{tableA:one-b}
\end{subtable}

\caption{Trajectory generation from single mode and multiple modes} \label{tableA:one}
\end{table}

\section{Trajectory Generation from The Most Prominent Mode}
We show in Table \ref{tableA:one} the ADE and FDE performance of our forecasting model when $K$ trajectories are generated from the most prominent mode only. In the table, \textbf{Ours}+\textbf{Multi} denotes the inference method that generates $K$ future trajectories from the $M$ modes. This method is the same as that described in the paper. \textbf{Ours}+\textbf{Single} denotes the inference method that generates $K$ future trajectories from the most prominent mode, which is identified by the weight distribution $\{w_{m}\}_{m=1}^{M}$. We can observe from the table that the best quality ($K \ge 12$) is degraded when the trajectories are generated from the most prominent mode only. On the other hand, \textbf{Ours}+\textbf{Single} shows nearly the same average quality performance as \textbf{Ours}+\textbf{Multi}. These are very natural results. When sampling a single future trajectory, the most prominent mode will be chosen for the sampling. Therefore, \textbf{Ours}+\textbf{Multi} and \textbf{Ours}+\textbf{Single} will show the same performance. On the other hand, when sampling multiple future trajectories, the trajectories generated by \textbf{Ours}+\textbf{Multi} will better reflect the true future trajectory distribution. Therefore, \textbf{Ours}+\textbf{Multi} will outperforms \textbf{Ours}+\textbf{Single} in terms of the best quality.

\section{Trajectory Generation Speed}
We ran our model on PC equipped with Intel i7, 32GB RAM, and a GPU (RTX 2080Ti). To generate 15 trajectories per vehicle, it takes around 0.02 sec. 

\section{Implementation Details}

\subsection{Candidate Lanes Acquisition}
We identify $M=10$ lane candidates for each target vehicle based on the method proposed in \cite{Kim, Narayanan, Chang}. The lane segments within the search radius (10 meters) from the current position of the vehicle are first found. Next, lane candidates 80 meters long in the vehicle's heading direction are obtained by attaching the preceding and succeeding lane segments based on lane connectivity information provided by the HD maps. The set of coordinate points for the lane candidates is re-sampled such that any two adjacent coordinate points have equal distance (1 meter). The ground-truth lane on which the target vehicle has moved during the future timesteps is identified by the Euclidean distance between the ground-truth future trajectory and the lane candidates. If the number of the identified lane candidates is less than $M$, we add fake lane candidates with coordinate points of (0, 0). If the number is greater than $M$, $M-1$ randomly selected lanes and the ground-truth lane are used.

\subsection{Details of Our Implementation}
\subsubsection{Preprocessing:}
Let $\mathbf{p}_{i}^{t}=(p_{x}^{t}, p_{y}^{t})$ denote the position of the vehicle $V_{i}$ at $t$. The speed $s$ (meter per second) and heading $h$ (radian) of the vehicle at $t$ are calculated as follows:
\begin{equation}
s = \psi \sqrt{(p_{x}^{t} - p_{x}^{t-1})^{2} + (p_{y}^{t} - p_{y}^{t-1})^{2}},
\end{equation}
\begin{equation}
h = \arctan{(\frac{p_{y}^{t} - p_{y}^{t-1}}{p_{x}^{t} - p_{x}^{t-1}})},
\end{equation}
where $\psi$ is the sampling rate. Let $\mathbf{l}_{f}^{m}$ denote the coordinate of the $f$-th point of the lane $\mathbf{L}^{m}$. The tangent vector $\mathbf{v}_{f}=(v_{f,x},v_{f,y})$ and its direction $d_{\mathbf{v}_{f}}$ at the point are calculated as follows:
\begin{equation}
\mathbf{v}_{f} = \mathbf{l}_{f}^{m} - \mathbf{l}_{f-1}^{m}
\end{equation}
\begin{equation}
d_{\mathbf{v}_{f}} = \arctan{(\frac{v_{f,y} - v_{f-1,y}}{v_{f,x} - v_{f-1,x}})}.
\end{equation}


\subsubsection{Feature Extraction Module:}
The positional data $\mathbf{X}_{i}$, $\mathbf{Y}_{i}$, and $\mathbf{L}^{m}$ are first preprocessed by the method proposed in this paper. Next, the data are embedded by single-layer MLPs followed by ReLU activation. The MLPs for $\mathbf{X}_{i}$ and $\mathbf{Y}_{i}$ take as input a 4-dimensional vector and output a 16-dimensional vector. The MLP for $\mathbf{L}^{m}$ takes as input a 5-dimensional vector and outputs a 64-dimensional vector. Finally, the embedded sequential vectors are encoded by LSTM networks. The final hidden states of the LSTM networks are used for the final encodings. The hidden state size of the LSTM networks for $\mathbf{X}_{i}$ and $\mathbf{Y}_{i}$ is 16. The hidden state size for $\mathbf{L}^{m}$ is 64.

\subsubsection{Scene Context Extraction Module:}
The attention operation between $\tilde{\mathbf{X}}_{i}$ and $\tilde{\mathbf{L}}^{(1:M)}$ for the context vector $\mathbf{a}_{i}^{m}$ is based on \cite{Bahdanau}. The context vector $\mathbf{b}_{i}^{m}$ is calculated as follows: The messages coming to the node $V_{i}$ are first calculated by a single-layer MLP followed by ReLU activation, which takes as input a 34-dimensional vector and outputs a 16-dimensional vector, and then summarized by the sum operation. The summarized message is used to update the hidden state of the node. To update the hidden state, we use a GRU cell, which takes as input a 16-dimensional vector and outputs a 16-dimensional hidden state vector. After the one round of the message passing, $\mathbf{b}_{i}^{m}$ is obtained by summing the hidden states of the neighboring nodes.

\subsubsection{Mode Selection Network:} 
Ten lane-level scene context vectors $\{\mathbf{c}_{i}^{m}\}$ are first embedded by a single-layer MLP followed by ReLU activation, which takes as input a 160-dimensional vector and outputs a 64-dimension vector. The embedded vectors are then concatenated and used as input to a single-layer MLP, which takes as input a 640-dimensional vector and outputs a 10-dimension vector, to obtain the latent vector $\mathbf{z}_{h}$.


\subsubsection{Encoder and Prior:}
The encoder produces the mean and variance vectors from the lane-level scene context vector $\mathbf{c}_{i}^{m}$ and the positional data encoding $\tilde{\mathbf{Y}}_{i}$. We use two two-layer MLPs for the mean and variance, respectively. The first layers of the MLPs take as input a 178-dimensional vector and output a 64-dimensional vector. The second layers take as input a 64-dimensional vector and output a 16-dimensional vector. The prior produces the mean and variance vectors from $\mathbf{c}_{i}^{m}$. The networks for the prior have the same structure as those for the encoder except that the first layers of the MLPs take as input a 160-dimensional vector. Finally note that we use ReLU activation for the first layers of the MLPs.

\subsubsection{Decoder:} 
To produce the next position $\hat{\mathbf{p}}_{i}^{t+1}$, the current position $\hat{\mathbf{p}}_{i}^{t}$ is first embedded by a single-layer MLP followed by ReLU activation, which takes as input a 2-dimensional vector and output a 16-dimensional vector. Next, $\mathbf{c}_{i}^{m}$, $\mathbf{z}_{l}$, and the embedding are concatenated and used as input to an LSTM network, which takes as input a 192-dimensional vector and outputs a 128-dimensional hidden state vector, to update the hidden state vector. The next position is obtained by a single-layer MLP, which takes as input a 128-dimensional vector and outputs a 2-dimensional vector.

\subsubsection{Discriminator:} 
The positional data $[\mathbf{Y}_{i};\Delta\mathbf{Y}_{i}]$ is first embedded by a single-layer MLP followed by ReLU activation, which takes as input a 4-dimensional vector and outputs a 16-dimensional vector. The embedded sequential data is then encoded by an LSTM network, which takes as input 16-dimensional sequential vectors and outputs 16-dimensional sequential hidden state vectors. The future encoding and lane encoding $\tilde{\mathbf{L}}_{m}$ are then used as input to a single-layer MLP to produce a scalar value. The MLP takes as input an 80-dimensional vector.

\subsubsection{Training:}
Adam optimizer \cite{Kingma15} is used for the optimization with initial learning rates of $10^{-4}$ (nuScenes) and $5 \times 10^{-4}$ (Argoverse Forecasting) and batch size of 8 for 100 (nuScenes) and 50 (Argoverse Forecasting) epochs. We evaluate the prediction performance after every three consecutive training epochs by using the validation samples in the training dataset. Whenever the prediction performance improves over the past, we save the model's network parameters. During the training, we use a cyclical annealing schedule \cite{Fu} for $\beta$.

\subsection{Details of Ablation Study}
We describe the details of the ablation study shown in section 4.3 of the paper. For $\mathbf{M1}$, we do not use the positional data preprocessing (PDP), VLI, V2I, and GAN regularization proposed in the paper. As a result, the lane-level scene context vector $\mathbf{c}_{i}^{m}$ is defined as $\mathbf{c}_{i}^{m} = [\tilde{\mathbf{X}}_{i}; \tilde{\mathbf{L}}^{m}]$. For $\mathbf{M3}$, we use the VLI so that $\mathbf{c}_{i}^{m} = [\tilde{\mathbf{X}}_{i}; \mathbf{a}_{i}^{m}]$. Finally, $\mathbf{c}_{i}^{m} = [\tilde{\mathbf{X}}_{i}; \mathbf{a}_{i}^{m}; \mathbf{b}_{i}^{m}]$ is used for $\mathbf{M4}$, which employ the VLI and V2I.

\subsection{Details of Baselines}
We describe the details of the baseline models shown in Figure 3 of the paper. For the figure, we exclude the scene context extraction module and discriminator to show how helpful the introduction of the hierarchical latent structure would be for the mitigation of mode blur. Finally, note that the trajectories depicted in Figure 3-(a) of the paper is generated from $\mathbf{M2}$.

\subsubsection{Baseline:} 
We train a generative model with a latent variable to model the trajectory distribution. One scene context vector $\mathbf{c}_{i}$ that condenses the information about all the modes of the distribution is first calculated as follows:
\begin{equation}
\mathbf{c}_{i} = [\tilde{\mathbf{X}}_{i}; \tilde{\mathbf{L}}^{ATT}],
\end{equation}
where $\tilde{\mathbf{L}}^{ATT}$ is the result of the attention operation \cite{Bahdanau} between $\tilde{\mathbf{X}}_{i}$ and $\tilde{\mathbf{L}}^{(1:M)}$. $\mathbf{c}_{i}$ is then used as input to the encoder, prior, and decoder. 

\subsubsection{Baseline+BOM:}
We train \textbf{Baseline} with the best-of-many (BOM) sample objective \cite{Bhattacharyya}. During the training, we let the model generate five trajectories per vehicle and select the trajectory with the minimum ADE out of the five for the $L2$-distance loss calculation.

\subsubsection{Baseline+NF:}
We train \textbf{Baseline} with normalizing flows (NF) \cite{Rezende}. We apply ten planar flow operations to a random vector that follows the normal distribution to obtain the final latent variable.

\end{document}